\documentclass{article}

\PassOptionsToPackage{numbers, compress}{natbib}
\usepackage[preprint]{neurips_2019}

\usepackage[utf8]{inputenc} 
\usepackage[T1]{fontenc}    
\usepackage{hyperref}       
\usepackage{url}            
\usepackage{booktabs}       
\usepackage{amsfonts}       
\usepackage{amsmath}
\usepackage{nicefrac}       
\usepackage{microtype}      
\usepackage{subcaption}

\usepackage{siunitx}

\usepackage{algorithm}
\usepackage{algorithmic}
\usepackage{letltxmacro}
\newlength{\commentindent}
\usepackage[]{todonotes}

\DeclareMathOperator*{\argmin}{arg\,min}

\title{Interactive Differentiable Simulation}

\author{%
  Eric Heiden\thanks{Equal contribution},\, David Millard${}^*$\!\!\!,\, Hejia Zhang, Gaurav S. Sukhatme \\[1em]
  University of Southern California, Los Angeles, USA\\[1em]
  \texttt{\{heiden,dmillard,hejiazha,gaurav\}@usc.edu}
}

\begin{document}

\maketitle

\begin{abstract}
  Intelligent agents need a physical understanding of the world to predict the impact of their actions in the future. While learning-based models of the environment dynamics have contributed to significant improvements in sample efficiency compared to model-free reinforcement learning algorithms, they typically fail to generalize to system states beyond the training data, while often grounding their predictions on non-interpretable latent variables.
  We introduce Interactive Differentiable Simulation (IDS), a differentiable physics engine, that allows for efficient, accurate inference of physical properties of rigid-body systems. Integrated into deep learning architectures, our model is able to accomplish system identification using visual input, leading to an interpretable model of the world whose parameters have physical meaning. We present experiments showing automatic task-based robot design and parameter estimation for nonlinear dynamical systems by automatically calculating gradients in IDS. When integrated into an adaptive model-predictive control algorithm, our approach exhibits orders of magnitude improvements in sample efficiency over model-free reinforcement learning algorithms on challenging nonlinear control domains.
\end{abstract}

\section{Introduction}

A key ingredient to achieving intelligent behavior is physical understanding. Under the umbrella of intuitive physics, specialized models, such as interaction and graph neural networks, have been proposed to learn dynamics from data to predict the motion of objects over long time horizons.  By labelling the training data given to these models by physical quantities, they are able to produce behavior that is conditioned on actual physical parameters, such as masses or friction coefficients, allowing for plausible estimation of physical properties and improved generalizability.

In this work, we introduce Interactive Differentiable Simulation (IDS), a differentiable physical simulator for rigid body dynamics. Instead of learning every aspect of such dynamics from data, our engine constrains the learning problem to the prediction of a small number of physical parameters that influence the motion and interaction of bodies.

A differentiable physics engine provides many advantages when used as part of a learning process. Physically accurate simulation obeys dynamical laws of real systems, including conservation of energy and momentum. Furthermore, joint constraints are enforced with no room outside of the model for error. The parameters of a physics engine are well-defined and correspond to properties of real systems, including multi-body geometries, masses, and inertia matrices. Learning these parameters provides a significantly interpretable parameter space, and can benefit classical control and estimation algorithms. Further, due to the high inductive bias, model parameters need not be jointly retrained for differing degrees of freedom or a reconfigured dynamics environment.

\section{Differentiable Rigid Body Dynamics}
\label{sec:physics}

\begin{figure}
    \centering
    \includegraphics[width=.6\textwidth]{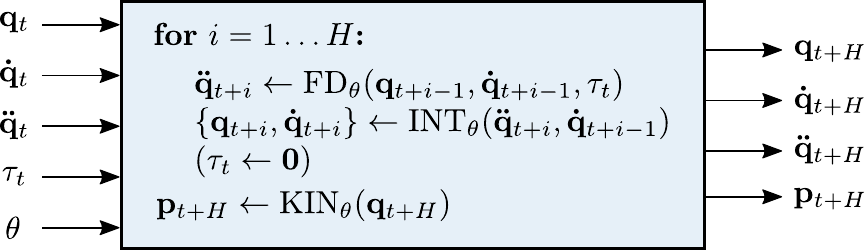}
    \caption{IDS deep learning layer with its possible inputs and outputs, unrolling our proposed dynamics model over $H$ time steps to compute future system quantities given current joint coordinates, joint forces $\mathbf{\tau}_t$ and model parameters $\theta$. $\operatorname{FD}(\cdot)$ computes the joint velocities, $\operatorname{INT}(\cdot)$ integrates accelerations and velocities, and $\operatorname{KIN}(\cdot)$ computes the 3D positions of all objects in the system in world coordinates. Depending on which quantities are of interest, only parts of the inputs and outputs are used. In some use cases, the joint forces are set to zero after the first computation to prevent their repeated application to the system. Being entirely differentiable, gradients of the motion of objects w.r.t the input parameters and forces are available to drive inference, design and control algorithms.}
    \label{fig:pytorch_autograd_function}
\end{figure}

In this work, we introduce a physical simulator for rigid-body dynamics. The motion of kinematic chains of multi-body systems can be described using the Newton-Euler equations:
\begin{align*}
\mathbf{M}(\mathbf{q})\mathbf{\ddot{q}} + \mathbf{C}(\mathbf{q}, \mathbf{\dot{q}}) = \mathbf{\tau}.
\end{align*}
Here, $\mathbf{q}$, $\mathbf{\dot{q}}$ and $\mathbf{\ddot{q}}$ are vectors of generalized\footnote{``Generalized coordinates'' sparsely encode only particular degrees of freedom in the kinematic chain such that bodies connected by joints are guaranteed to remain connected.} position, velocity and acceleration coordinates, and $\mathbf{\tau}$ is a vector of generalized forces. $\mathbf{M}$ is the generalized inertia matrix and depends on $\mathbf{q}$. Coriolis forces, centrifugal forces, gravity and other forces acting on the system, are accounted for by the bias force matrix $\mathbf{C}$ that depends on $\mathbf{q}$ and $\mathbf{\dot{q}}$. Since all bodies are connected via joints, including free-floating bodies which connect to a static world body via special joints with seven degrees of freedom (DOF), i.e. 3D position and orientation in quaternion coordinates, their positions and orientations in world coordinates $\mathbf{p}$ are computed by the forward kinematics function $\operatorname{KIN}(\cdot)$ (Fig.~\ref{fig:pytorch_autograd_function}) using the joint angles and the bodies' relative transforms to their respective joints through which they are attached to their parent body.

\begin{figure}
    \centering
    \includegraphics[trim=12cm 0cm 12cm 0cm,clip,width=.23\linewidth]{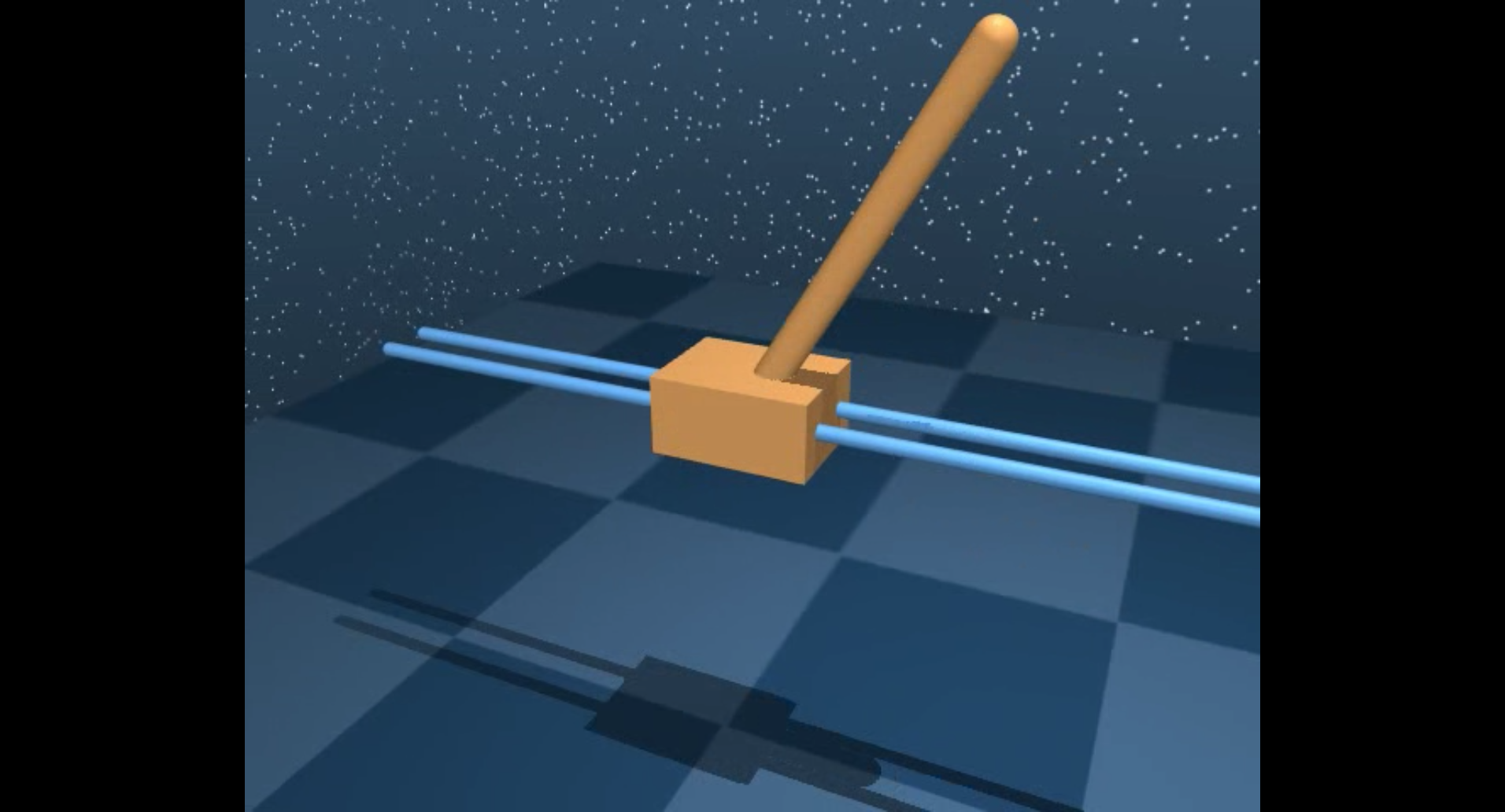}
    \includegraphics[trim=15cm 2cm 15cm 0cm,clip,width=.23\linewidth]{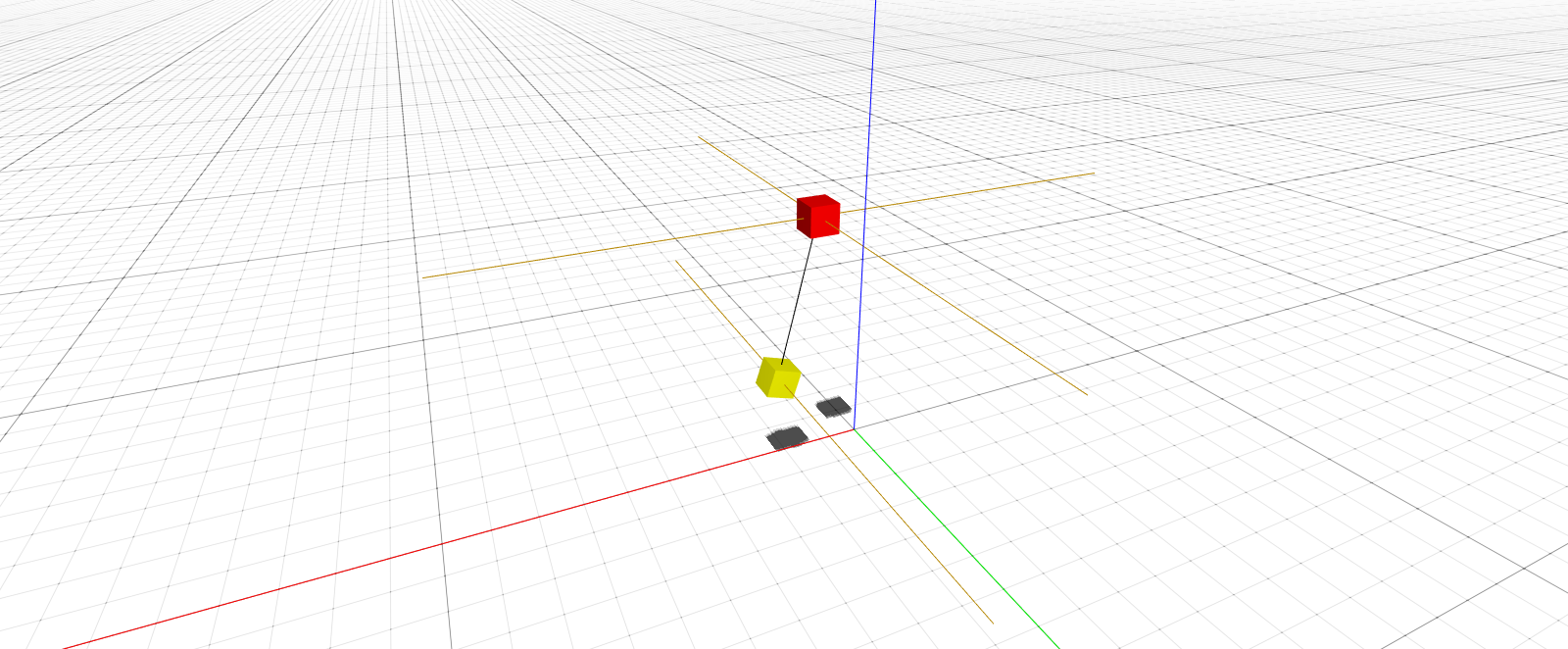}\hfill
    \includegraphics[trim=4cm 3cm 4cm 7cm,clip,width=.23\linewidth]{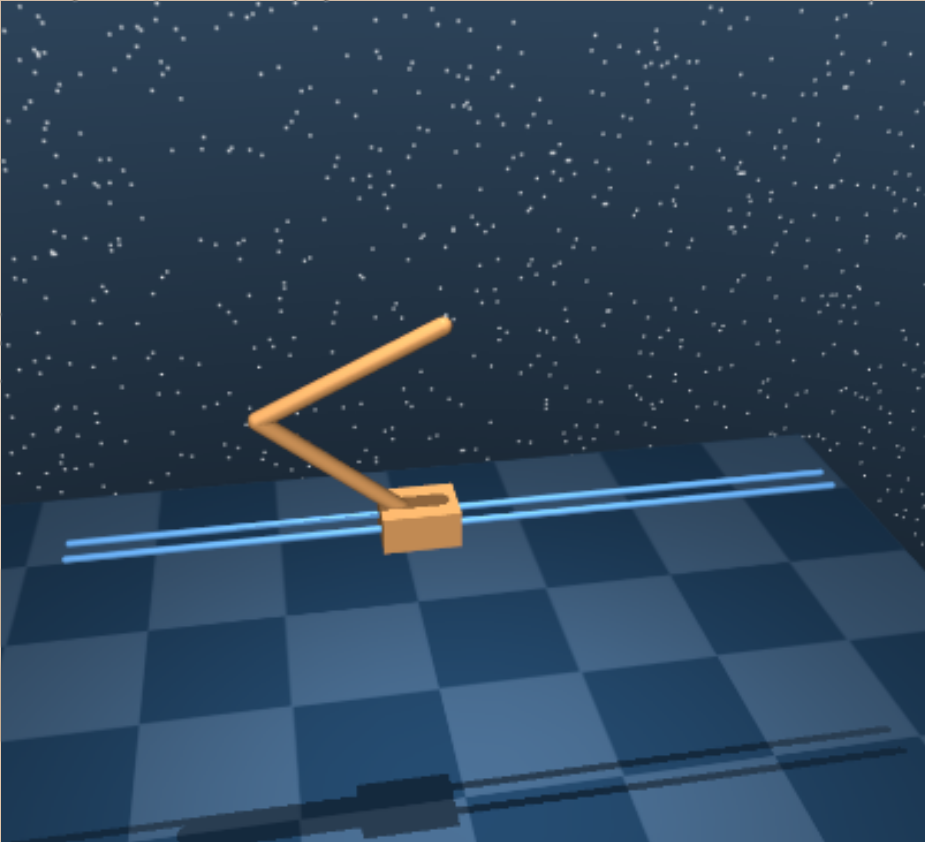}
    \includegraphics[trim=15cm 2cm 15cm 4cm,clip,width=.23\linewidth]{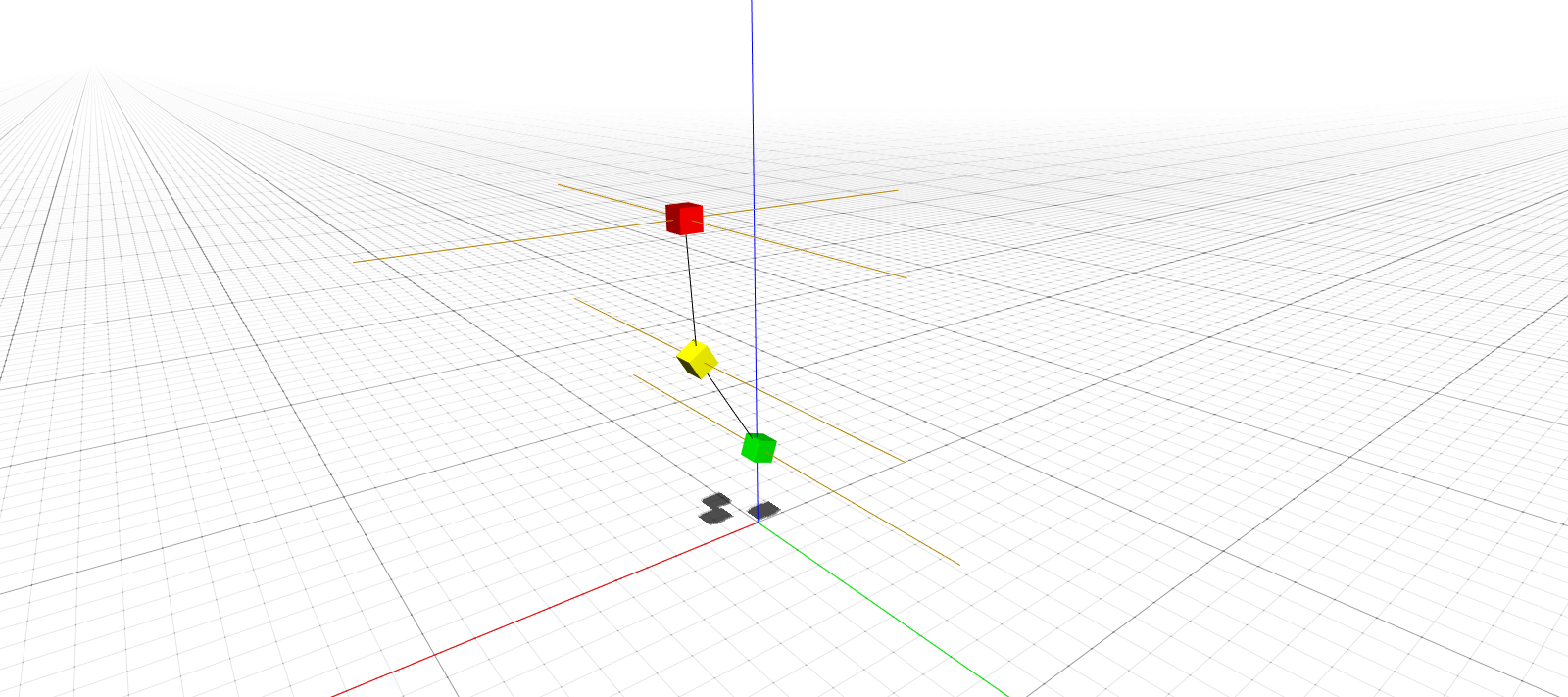}
    \caption{Visualization of the physical models and environments used. Single cartpole environment (left). Double cartpole environment (right). Both are actuated by a linear force applied to the cart. The cart is constrained to the rail, but may move infinitely in either direction. Blue backgrounds show environments from DeepMind Control Suite~\cite{tassa2018dm} in the MuJoCo~\cite{todorov2012mujoco} physics simulator. Visualizations with white background show Interactive Differentiable Simulation (IDS), our approach.}
    \label{fig:models}
\end{figure}

Forward dynamics $\operatorname{FD}(\cdot)$ (cf. Fig.~\ref{fig:pytorch_autograd_function}) is the mapping from positions, velocities and forces to accelerations. We efficiently compute the forward dynamics using the Articulated Body Algorithm (ABA)~\cite{featherstone2007rbda}. Given a descriptive model consisting of joints, bodies, and predecessor/successor relationships, we build a kinematic chain that specifies the dynamics of the system. In our simulator, bodies comprise physical entities with mass, inertia, and attached rendering and collision geometries. Joints describe constraints on the relative motion of bodies in a model.  Equipped with such a graph of $n$ bodies connected via joints with forces acting on them, ABA computes the joint accelerations $\mathbf{\ddot{q}}$ in $O(n)$ operations. Following the calculation of the accelerations, we implement semi-implicit Euler integration (referred to as $\operatorname{INT}(\cdot)$ in Fig.~\ref{fig:pytorch_autograd_function}) to compute the velocities and positions of the joints and bodies at the current instant $t$ given time step $\Delta_t$:
\begin{align*}
    \mathbf{\dot{q}}_{t} &= \mathbf{\dot{q}}_{t-1} + \Delta_t\mathbf{\ddot{q}}_{t} &
    \mathbf{q}_{t} &= \mathbf{q}_{t-1} + \Delta_t\mathbf{\dot{q}}_{t}
\end{align*}
In control scenarios, external forces are applied to the kinematic tree, which are then propagated through the joints and bodies of the physical model. This propagation is efficiently calculated using the Recursive Newton-Euler Algorithm~\cite{featherstone2007rbda}. For body $i$, let $\lambda(i)$ denote the predecessor body and $\mu(i)$ denote successor bodies. We denote the allowable motion subspace matrix of the joint by $S_i$, and the spatial inertia matrix by $I_i$. Given the velocity $v_i$ and acceleration $a_i$ of body $i$, we may compute
\begin{align*}
v_i &= v_{\lambda(i)} + S_i \dot{q}_i &
a_i &= a_{\lambda(i)} + S_i \ddot{q}_i + \dot{S}_i \dot{q}_i.
\end{align*}
Denoting the net force on body $i$ as $f_i^B$, we use the physical equation of motion to relate this force to the body acceleration
\begin{align*}
f_i^B = I_i a_i + v_i \times^* I_i v_i.
\end{align*}
We can separate this force into $f_i$, the force transmitted from body $\lambda(i)$ across joint $i$, and $f_i^x$, the external force acting on body $i$ (such as gravity). Then
\begin{align*}
f_i^B = f_i + f_i^x - \sum_{j \in \mu(i)} f_j,
\end{align*}
which lets us easily calculate $f_i$, the force transmitted across each joint as
\begin{align*}
f_i = f_i^B - f_i^x + \sum_{j \in \mu(i)} f_j.
\end{align*}
Finally, we may calculate the generalized force vector $\tau_i$ at joint $i$ as
\begin{align*}
\tau_i = S_i^{\mathrm{T}} f_i.
\end{align*}

While the analytical gradients of the rigid-body dynamics algorithms can be derived manually~\cite{carpentier2018analytical}, we choose to implement the entire physics engine in the reverse-mode automatic differentiation framework Stan Math~\cite{carpenter2015stan}. Automatic differentiation allows us to compute gradients of any quantity involved in the simulation of complex systems, opening avenues to state estimation, optimal control and system design. Enabled by low-level optimization, our C++ implementation is designed to lay the foundations for real-time performance on physical robotic systems in the future.

\section{Experiments}
\label{sec:experiments}

\subsection{Inferring Physical Properties from Vision}
\label{sec:exp_vision}

To act autonomously, intelligent agents need to understand the world through high-dimensional sensor inputs, like cameras. We demonstrate that our approach is able to infer the relevant physical parameters of the environment dynamics from these types of high-dimensional observations. We optimize the weights of an autoencoder network trained to predict the future visual state of a dynamical system, with our physics layer serving as the bottleneck layer. In this exemplar scenario, given an image of a three-link compound pendulum simulated in the MuJoCo physics simulator~\cite{todorov2012mujoco} at time $t$, the model is tasked to predict the future rendering of this pendulum $H$ time steps ahead. Compound pendula are known to exhibit chaotic behavior, i.e. given slightly different initial conditions (such as link lengths, starting angles, etc.), the trajectories drift apart significantly. Therefore, IDS must recover the true physical parameters accurately in order to generate valid motions that match the training data well into the future.

We model the encoder $f_{\mathrm{enc}}$ and the decoder $f_{\mathrm{dec}}$ as neural networks consisting of two 256-unit hidden layers mapping from $100\times100$ grayscale images to a six-dimensional vector of joint positions $\mathbf{q}$ and velocities $\mathbf{\dot{q}}$, and vice-versa. Inserted between both networks, we place an IDS layer (Fig.~\ref{fig:pytorch_autograd_function}) to forward-simulate the given joint coordinates from time $t$ to time $t+H$, where $H$ is the number of time steps of the prediction horizon. Given that the input data uses a time step $\Delta_t=\SI{0.05}{\second}$, the goal is to predict the state of the pendulum $\SI{1}{\second}$ into the future. While the linear layers of $f_{\mathrm{enc}}$ and $f_{\mathrm{dec}}$ are parameterized by weights and biases, IDS, referred to as $f_{\mathrm{phy}}$, is conditioned on physical parameters $\theta_{\mathrm{phy}}$ which, in the case of our compound pendulum, are the lengths of the three links $\{l_0, l_1, l_2\}$. We choose arbitrary values $\{1, 5, 0.5\}$ to initialize these parameters.

\begin{figure}
    \centering
    \begin{subfigure}[t]{.5\linewidth}
    \includegraphics[height=4cm]{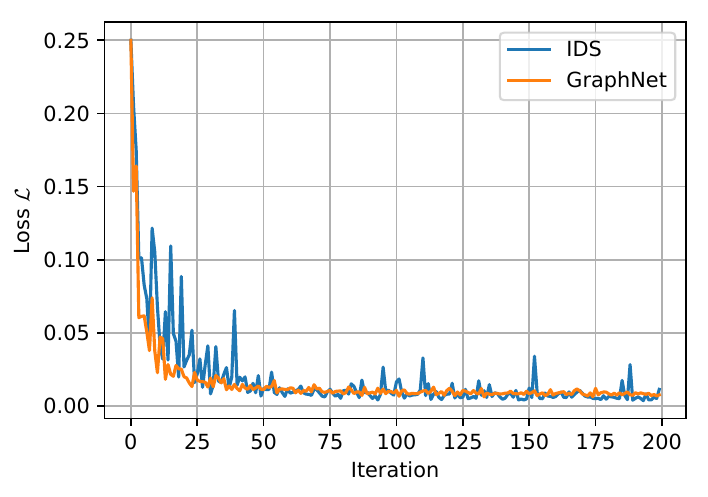}
    \end{subfigure}
    \hspace{0.05\linewidth}
    \begin{subfigure}[t]{.4\linewidth}
    \includegraphics[height=4cm]{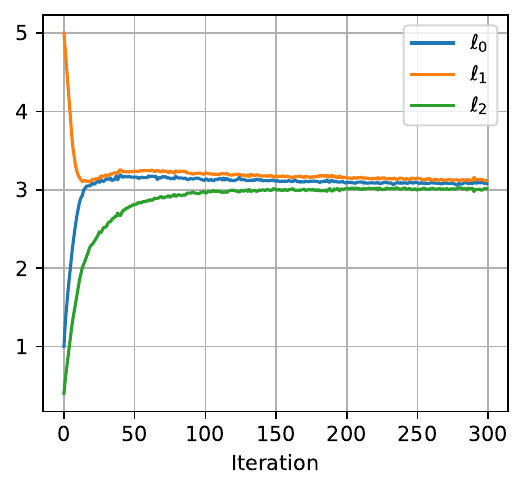}
    \end{subfigure}
    \caption{(left) Learning curve of the triple loss $\mathcal{L}$ (Eq.~\eqref{eq:bigmac_loss}). An autoencoder trained to predict the future using our physics module trains at a comparable rate to the state-of-the-art intuitive physics approach based on graph neural networks~\cite{hadsell2018graphnet} as the predictive bottleneck. (right) Integrated as the bottleneck in a predictive autoencoder, IDS accurately infers the model parameters $\theta_\mathrm{phy}$ of a compound pendulum, which represent the three link lengths.}
    \label{fig:bigmac_training}
\end{figure}

Given a dataset $D$ of ground-truth pairs of images $(o_{t}^*, o_{t+H}^*)$ and ground-truth joint coordinates $(\mathbf{q}_t^*, \mathbf{q}_{t+1}^*, \mathbf{\dot{q}}_t^*, \mathbf{\dot{q}}_{t+1}^*)$, we optimize a triplet loss using the Adam optimizer that jointly trains the individual components of the autoencoder:
\begin{align}
\nonumber
\mathcal{L}_{\mathrm{enc}} &= || f_{\mathrm{enc}}(o_t) - [\mathbf{q}_{t}^*, \mathbf{\dot{q}}_{t}^*]^T ||_2^2 \\
\nonumber
\mathcal{L}_{\mathrm{phy}} &= || f_{\mathrm{phy}}([\mathbf{q}_t, \mathbf{\dot{q}}_t]) - [\mathbf{q}_{t+H}^*, \mathbf{\dot{q}}_{t+H}^*]^T ||_2^2 \\
\nonumber
\mathcal{L}_{\mathrm{dec}} &= || f_{\mathrm{dec}}([\mathbf{q}_{t+H}, \mathbf{\dot{q}}_{t+H}]) - o_{t+H}^* ||_2^2 \\
\mathcal{L}(\cdot; \theta_{\mathrm{enc}}, \theta_{\mathrm{phy}}, \theta_{\mathrm{dec}}) &=
\sum_D 
\mathcal{L}_\mathrm{enc}(\cdot; \theta_{\mathrm{enc}}) +
\mathcal{L}_\mathrm{phy}(\cdot; \theta_{\mathrm{phy}}) +
\mathcal{L}_\mathrm{dec}(\cdot; \theta_{\mathrm{dec}})
\label{eq:bigmac_loss}
\end{align}
We note that the physical parameters $\theta_{\mathrm{phy}}$ converge to the true parameters of the dynamical system ($l_0=l_1=l_2=3$), as shown in Fig.~\ref{fig:bigmac_training}.

As a baseline from the intuitive physics literature, we train a graph neural network model based on~\cite{hadsell2018graphnet} on the first 800 frames of a 3-link pendulum motion. When we let the graph network predict 20 time steps into the future from a point after these 800 training samples, it returns false predictions where the pendulum is continuing to swing up, even though such motion would violate Newton's laws. Such behavior is typical for fully learned models, which mostly achieve accurate predictions within the domain of the training examples. By contrast, IDS imposes a strong inductive bias, which allows the estimator to make accurate predictions far into the future (Fig.~\ref{fig:bigmac_training}).

\begin{figure}
    \centering
    \includegraphics[width=.7\linewidth]{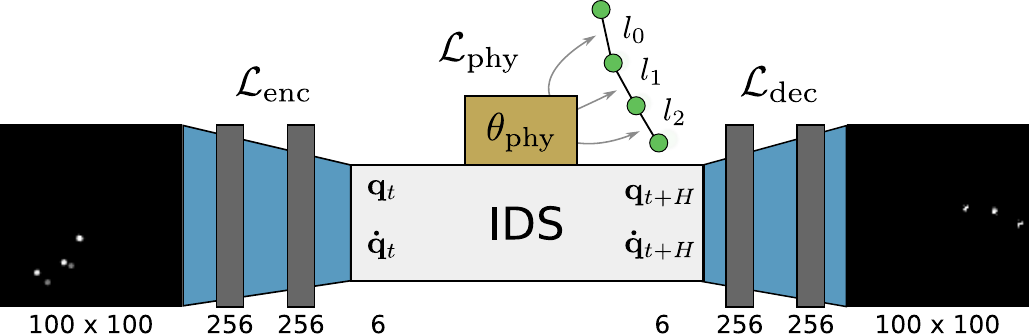}
    \caption{Architecture of the autoencoder encoding grayscale images of a three-link pendulum simulated in MuJoCo to joint positions and velocities, $\mathbf{q}$, $\mathbf{\dot{q}}$, respectively, advancing the state of the system by $H$ time steps and producing an output image of the future state the system. The parameters of the encoder, decoder and our physics layer are optimized jointly to minimize the triplet loss from Eq.~\ref{eq:bigmac_loss}.}
    \label{fig:autoencoder}
\end{figure}

\subsection{Automatic Robot Design}
Industrial robotic applications often require a robot to follow a given tool path. In general, robotic arms with 6 or more degrees of freedom provide large workspaces and redundant configurations to reach any possible point within the workspace. However, motors are expensive to produce, maintain, and calibrate. Designing arms that contain a minimal number of motors required for a task provides economic and reliability benefits, but imposes constraints on the kinematic design of the arm.

One standard for specifying the kinematic configuration of a serial robot arm is the Denavit-Hartenberg (DH) parameterization. For each joint $i$, the DH parameters are $(d_i, \theta_i, a_i, \alpha_i)$. The preceding motor axis is denoted by $z_{i-1}$ and the current motor axis is denoted by $z_i$. $d_i$ describes the distance to the joint $i$ projected onto $z_{i-1}$ and $\theta_i$ specifies the angle of rotation about $z_{i-1}$. $a_i$ specifies the distance to joint $i$ in the direction orthogonal to $z_{i-1}$ and $\alpha_i$ describes the angle between $z_i$ and $z_{i-1}$, rotated about the $x$-axis of the preceding motor coordinate frame. We are primarily interested in arms with motorized revolute joints, and thus $\theta_i$ becomes the $q_i$ parameter of our joint state. We can thus fully specify the relevant kinematic properties of a serial robot arm with $N$ degrees of freedom (DOF) as $R = \{ d_0, a_0, \alpha_0, \ldots, d_N, a_N, \alpha_N \}$.

\begin{figure}
    \centering
    \begin{subfigure}[t]{.65\linewidth}
    \includegraphics[width=\linewidth]{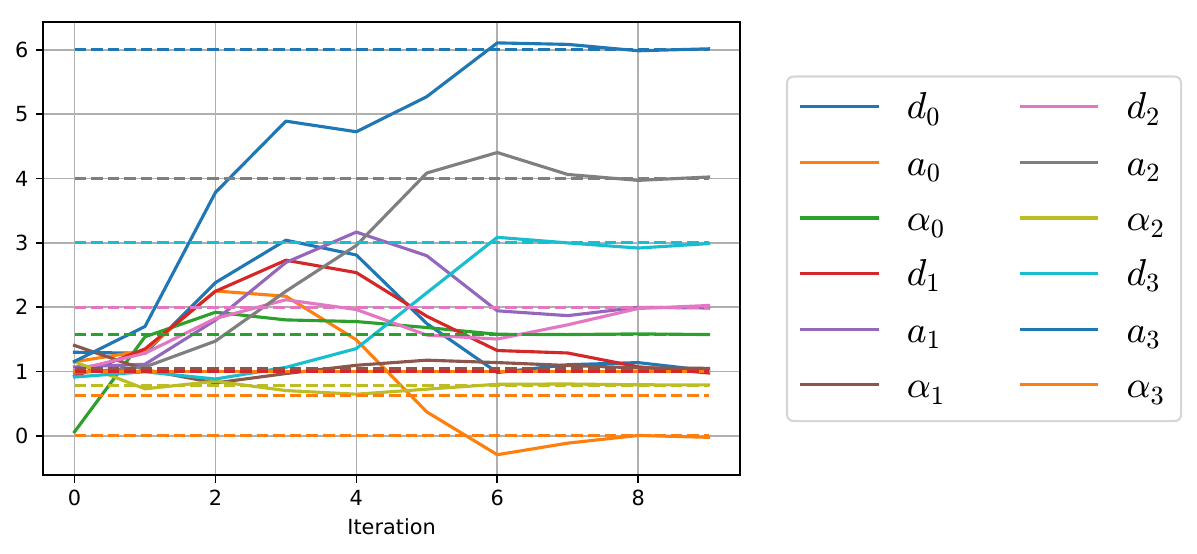}
    \end{subfigure}
    \hfill
    \begin{subfigure}[t]{.28\linewidth}
    \includegraphics[width=\linewidth,trim=22cm 1cm 14cm 3cm,clip]{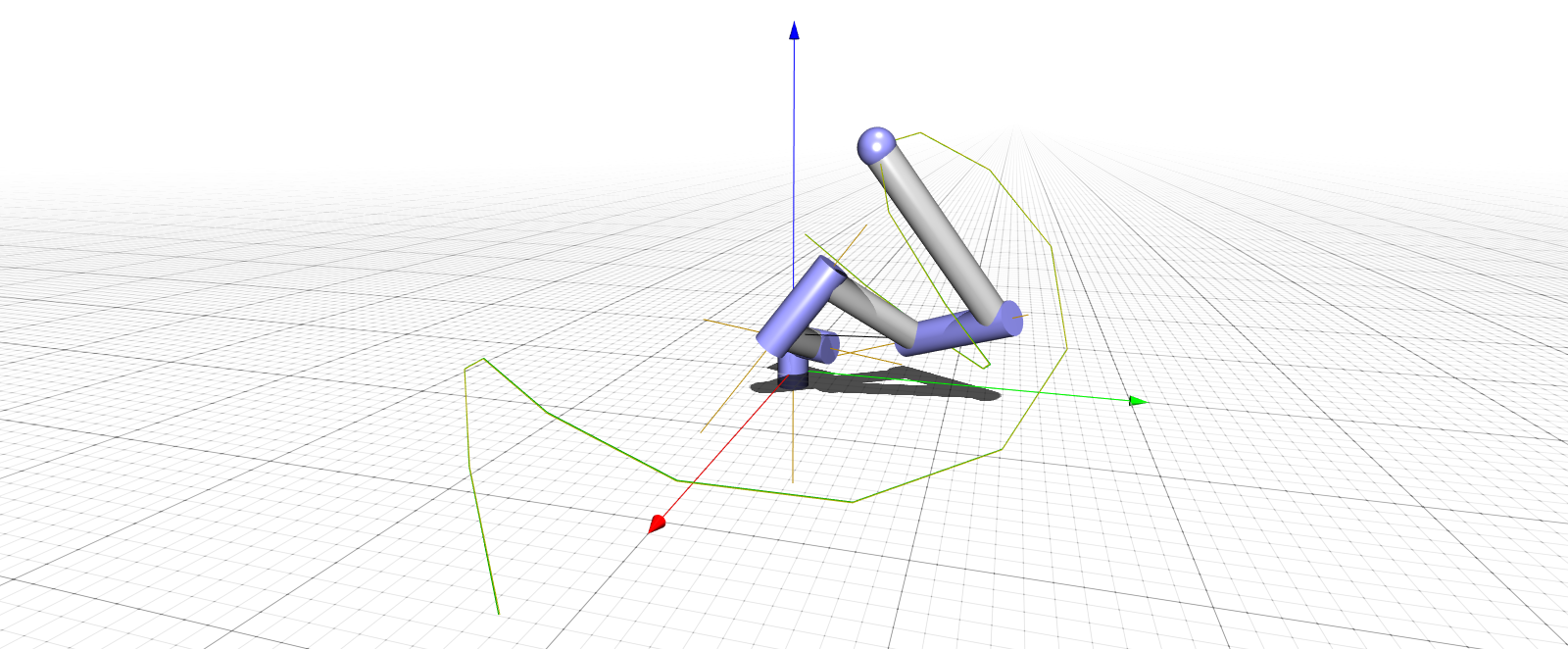}
    \end{subfigure}
    \caption{(left) Using the L-BFGS optimizer equipped with gradients through the kinematics equations of IDS, we are able to optimize the Denavit-Hartenberg (DH) parameters of a 4-DOF robot arm to those of the robot arm used to generate a feasible trajectory. (right) Visualization of a 4-DOF robot arm and its trajectory in IDS. }\label{fig:dh_params}
\end{figure}

We specify a task-space trajectory $\tau = \{ p_0, p_1, \ldots, p_T \}$ for $p_t \in \mathbb{R}^3$ as the world coordinates of the end-effector of the robot. Given a joint-space trajectory $\{ q_0, q_1, \ldots, q_T \}$, we seek to find the best $N$-DOF robot arm design, parameterized by DH vector $R$, that most closely matches the specified end-effector trajectory:
\begin{align*}
    R^* = \argmin_R \sum_{t=0}^{T} ||\operatorname{KIN}(q_t ; R) - p_t||_2^2,
\end{align*}
where the forward kinematics function $\operatorname{KIN}(\cdot)$ maps from joint space to Cartesian tool positions conditioned on DH parameters $R$. Since we compute $\operatorname{KIN}(\cdot)$ using our engine, we may compute derivatives to arbitrary inputs to this function (cf. Fig.~\ref{fig:pytorch_autograd_function}), and use gradient-based optimization through L-BFGS to converge to arm designs which accurately perform the trajectory-following task, as shown in Fig.~\ref{fig:dh_params}.

\subsection{Adaptive MPC}
\label{sec:ampc}
Besides parameter estimation and design, a key benefit of differentiable physics is its applicability to optimal control algorithms. In order to control a system within our simulator, we specify the control space $\mathbf{u}$, which is typically a subset of the system's generalized forces $\tau$, and the state space $\mathbf{x}$. Given a quadratic, i.e. twice-differentiable, cost function \mbox{$c:\ \mathbf{x}\times\mathbf{u}\to\mathbb{R}$}, we can linearize the dynamics at every time step, allowing efficient gradient-based optimal control techniques to be employed. Iterative Linear Quadratic Control~\cite{li2004iterative} (iLQR) is a direct trajectory optimization algorithm that uses a dynamic programming scheme on the linearized dynamics to derive the control inputs that successively move the trajectory of states and controls closer to the optimum of the cost function.

Throughout our control experiments, we optimize a trajectory for an $n$-link cartpole to swing up from arbitrary initial configuration of the joint angles. In the case of double cartpole, i.e. a double inverted pendulum on a cart, the state space is defined as
$
\mathbf{x} = \left(p, \dot{p}, \sin q_0, \cos q_0, \sin q_1, \cos q_1, \dot{q}_0, \dot{q}_1, \ddot{q}_0, \ddot{q}_1\right),
$
where $p$ and $\dot{p}$ refer to the cart's position and velocity, $q_0, q_1$ to the joint angles, and $\dot{q}_0, \dot{q}_1, \ddot{q}_0, \ddot{q}_1$ to the velocities and accelerations of the revolute joints of the poles, respectively. For a single cartpole the state space is analogously represented, excluding the second revolute joint coordinates $q_1,\dot{q}_1,\ddot{q}_1$. The cost is defined as the norm of the control plus the Euclidean distance between the cartpole's current state and the goal state
$
\mathbf{x}^* = \left(0, 0, 0, 1, 0, 1, 0, 0, 0, 0\right),
$
at which the pole is upright at zero angular velocity and acceleration, and the cart is centered at the origin with zero positional velocity.

Trajectory optimization assumes that the dynamics model is accurate w.r.t the real world and generates sequences of actions that achieve optimal behavior towards a given goal state, leading to open-loop control. Model-predictive control (MPC) leverages trajectory optimization in a feedback loop where the next action is chosen as the first control computed by trajectory optimization over a shorter time horizon with the internal dynamics model. After some actions are executed in the real world and subsequent state samples are observed, \emph{adaptive} MPC (Algorithm~\ref{algo:ampc}) fits the dynamics model to these samples to align it closer with the real-world dynamics. In this experiment, we want to investigate how differentiable physics can help overcome the domain shift that poses an essential challenge of model-based control algorithms that are employed in a different environment. To this end, we incorporate IDS as dynamics model in such receding-horizon control algorithm to achieve swing-up motions of a single and double cartpole in the DeepMind Control Suite~\cite{tassa2018dm} environments that are based on the MuJoCo simulator~\cite{todorov2012mujoco}.

\begin{figure}
    \centering
    \includegraphics[width=0.65\textwidth]{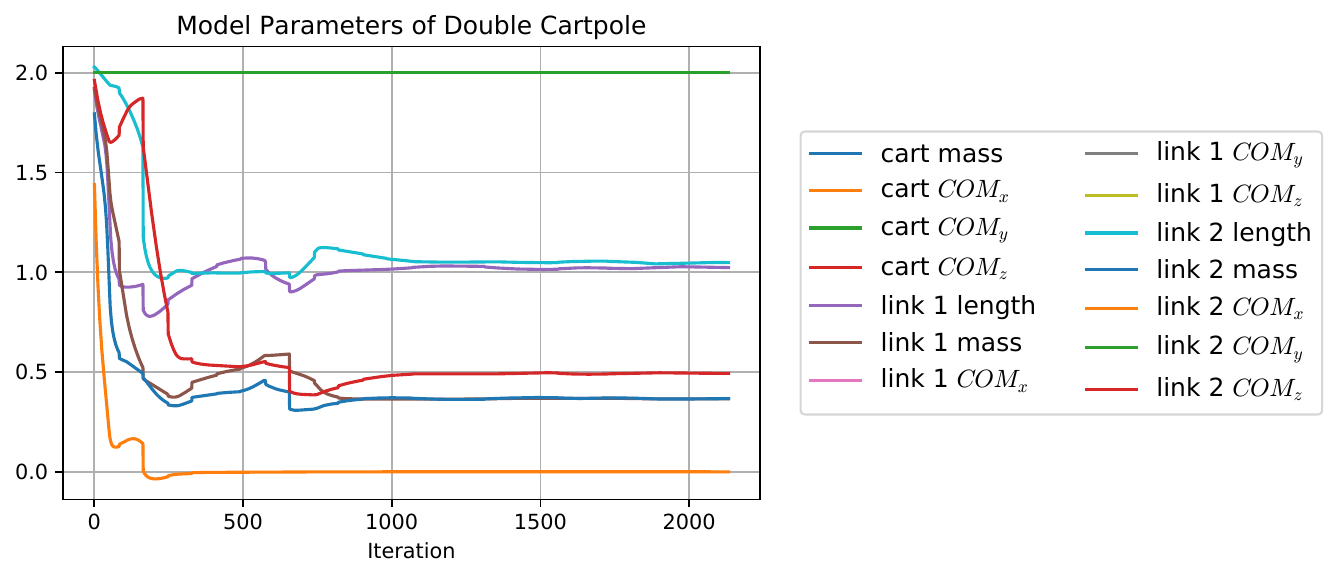}
    \caption{Convergence of the physical parameters of a double cartpole, over all model fitting iterations combined, using Adaptive MPC (Algorithm~\ref{algo:ampc}) in the DeepMind Control Suite environment.}
    \label{fig:ampc_cartpole_param}
\end{figure}

\begin{algorithm}
\begin{algorithmic}
\REQUIRE Cost function $c: \mathbf{x}\times\mathbf{u}\to\mathbb{R}$
\FOR{episode = $1..M$}
    \STATE $R\gets \emptyset$
    \qquad\COMMENT{Replay buffer to store transition samples from the real environment}
    \STATE Obtain initial state $x_{0}$ from the real environment
    \FOR{$t=1..T$}
        \STATE $\displaystyle \{u^*\}_{t}^{t+H} \gets \argmin_{u_{1:H}} \sum_{i=1}^H c(x_i, u_i)$
        \COMMENT{Trajectory optimization using iLQR}
        \STATE \qquad\qquad\qquad s.t. $x_1 = x_{t-1}$, $x_{i+1}=f_\theta(x_i, u_i)$, $\underbar{$u$}\leq u \leq \bar{u}$
        \STATE Take action $u_t$ in the real environment and obtain next state $x_{t+1}$
        \STATE Store transition $(x_t, u_t, x_{t+1})$ in $R$
    \ENDFOR
    \STATE Fit dynamics model $f_\theta$ to $R$ by minimizing the state-action prediction loss (Eq.~\eqref{eq:xu_loss})
\ENDFOR
\end{algorithmic}
\caption{Adaptive MPC algorithm using differentiable physics.}
\label{algo:ampc}
\end{algorithm}

We fit the parameters $\theta$ of the system
by minimizing the state-action prediction loss:
\begin{equation}
    \theta^* = \argmin_{\theta} \sum_{t} ||f_\theta (x_t, u_t) - x_{t+1}||_2^2
\label{eq:xu_loss}
\end{equation}
Thanks to the low dimensionality of the model parameter vector $\theta$ (for a double cartpole there are 14 parameters, cf. Fig.~\ref{fig:ampc_cartpole_param}), efficient optimizers such as the quasi-Newton optimizer L-BFGS are applicable, leading to fast convergence of the fitting phase, typically within 10 optimization steps. The length $T$ of one episode is 140 time steps. During the first episode we fit the dynamics model more often, i.e. every 50 time steps, to warm-start the receding-horizon control scheme. Given a horizon size $H$ of 20 and 40 time steps, MPC is able to find the optimal swing-up trajectory for the single and double cartpole, respectively.

\begin{figure}
    \centering
    \includegraphics[width=.48\textwidth,trim=0 8cm 0 0,clip]{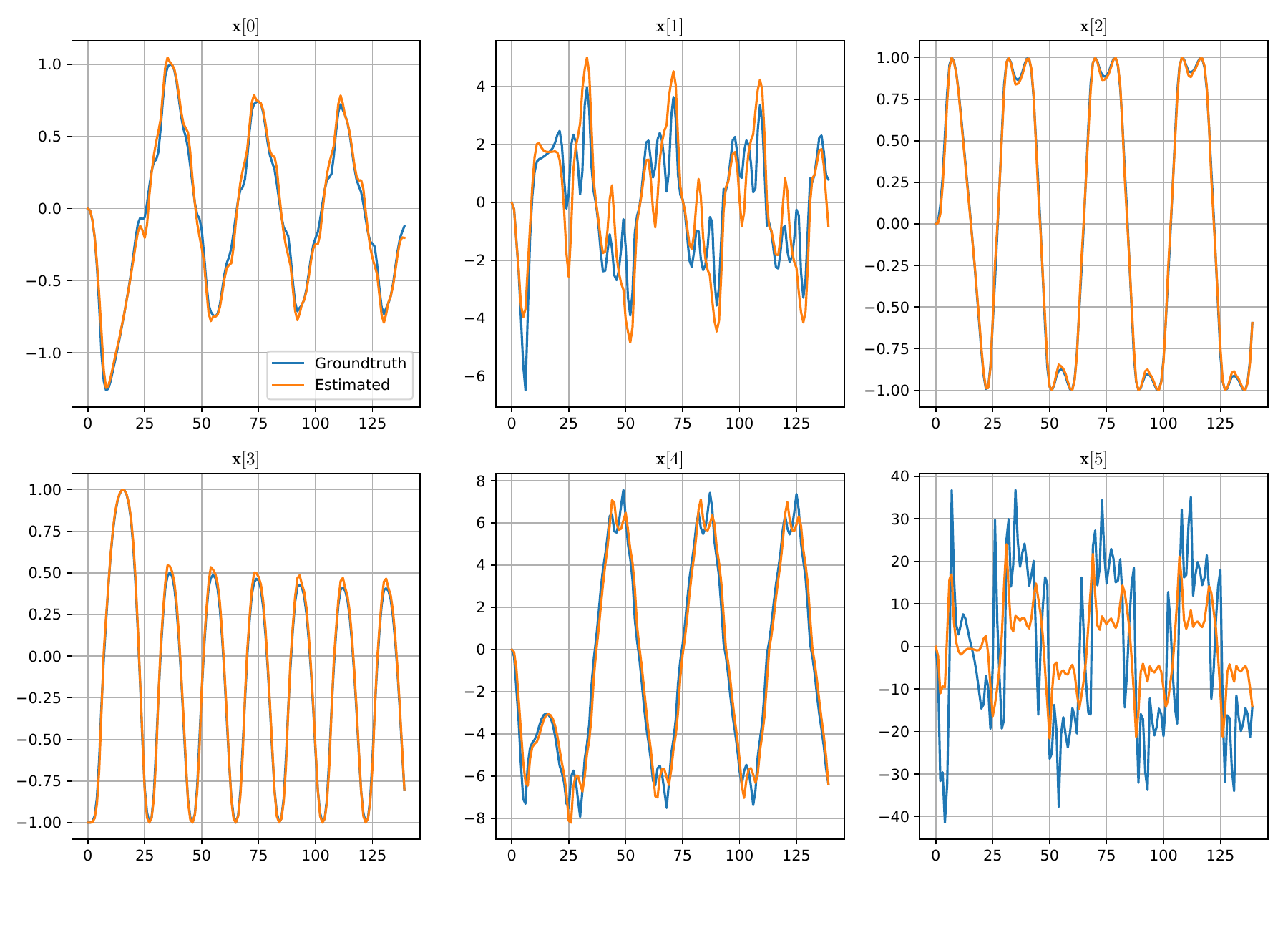}\hfill
    \includegraphics[width=.48\textwidth,trim=0 8cm 0 0,clip]{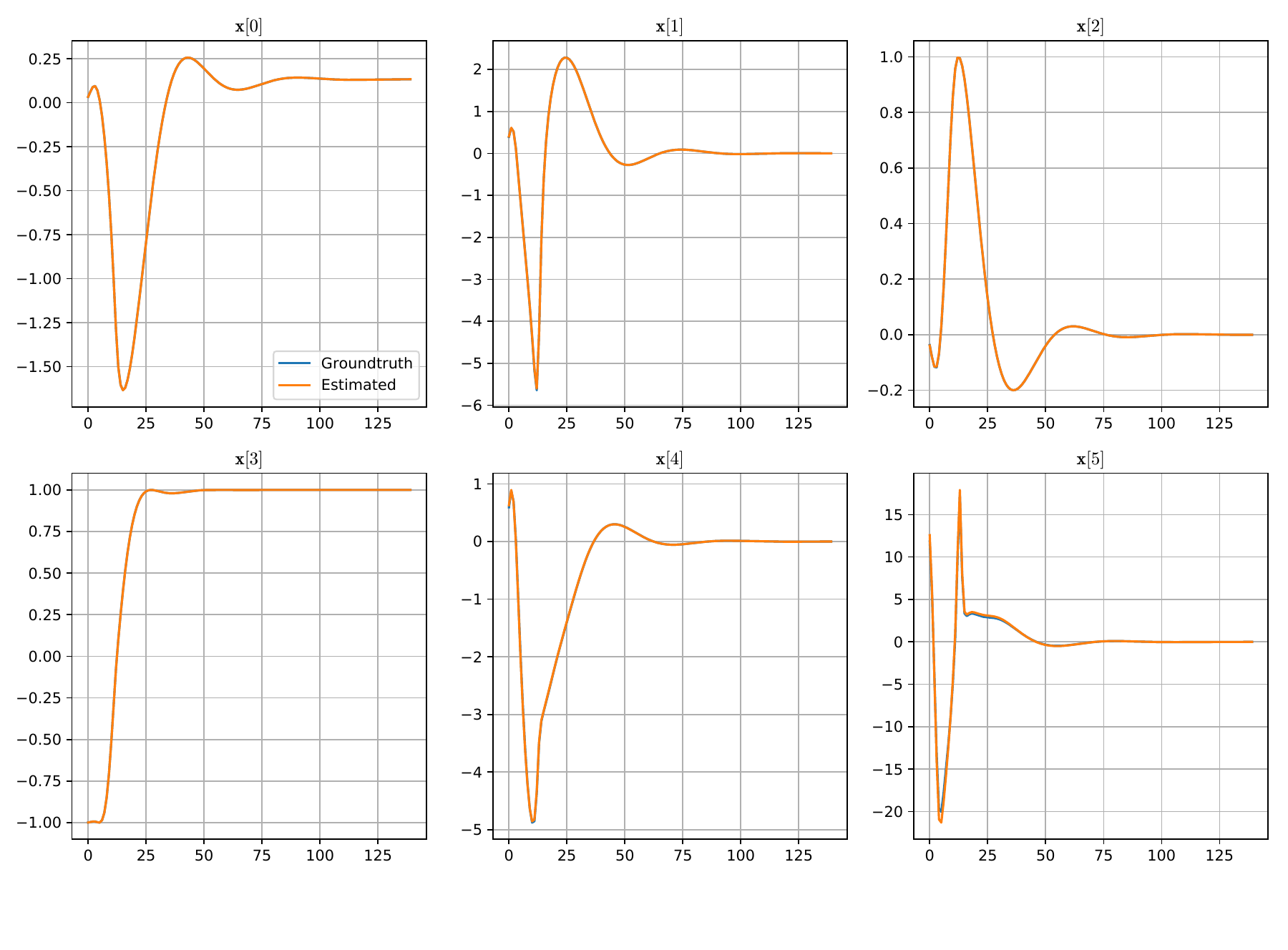}
    \caption{Trajectory of the six-dimensional cartpole state over 140 time steps after the first (left) and third (right) episode of Adaptive MPC (Algorithm \ref{algo:ampc}). After two episodes, the differentiable physics model (orange) has converged to model parameters that allow it to accurately predict the cartpole dynamics modelled in MuJoCo (blue). Since by the third episode the control algorithm has converged to a successfull cartpole swing-up, the trajectory differ significantly with the first roll-out.}
    \label{fig:rl_mpc_fitting}
\end{figure}

Within a handful of training episodes, adaptive MPC infers the correct model parameters involved in the dynamics of double cartpole (Fig.~\ref{fig:ampc_cartpole_param}). As shown in Fig.~\ref{fig:models}, the models we start from in IDS do not match their counterparts from DeepMind Control Suite. For example, the poles are represented by capsules where the mass is distributed across these elongated geometries, whereas initially in our IDS model, the center of mass of the poles is at the end of them, such that they have different inertia parameters. We set the masses, lengths of the links, and 3D coordinates of the center of masses to 2, and using a few steps of the optimizer and less than 100 transition samples, converge to a much more accurate model of the true dynamics in the MuJoCo environment. On the example of a cartpole, Fig.~\ref{fig:rl_mpc_fitting} visualizes the predicted and actual dynamics for each state dimension after the first (left) and third (right) episode.

\begin{figure}
    \centering
    \includegraphics[height=4cm]{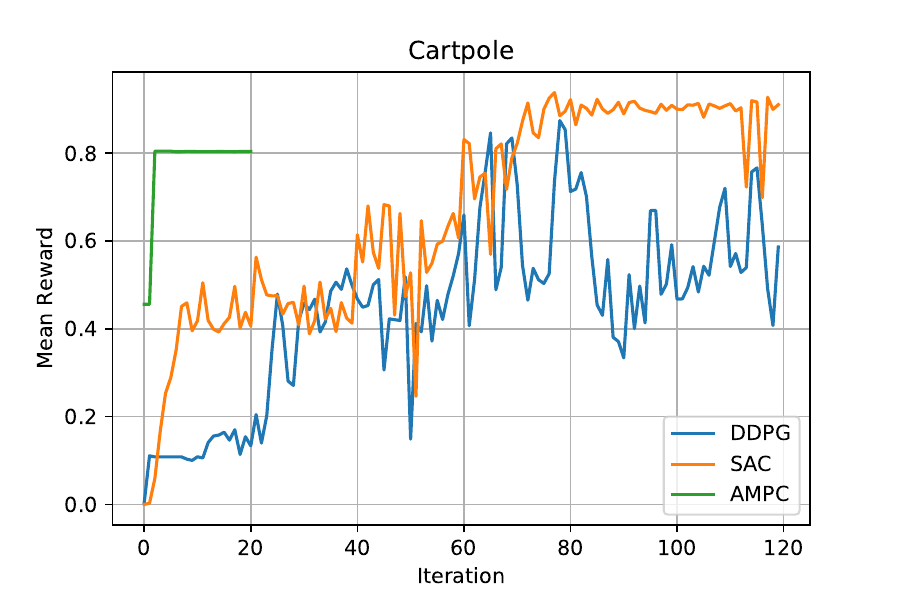}
    \includegraphics[height=4cm]{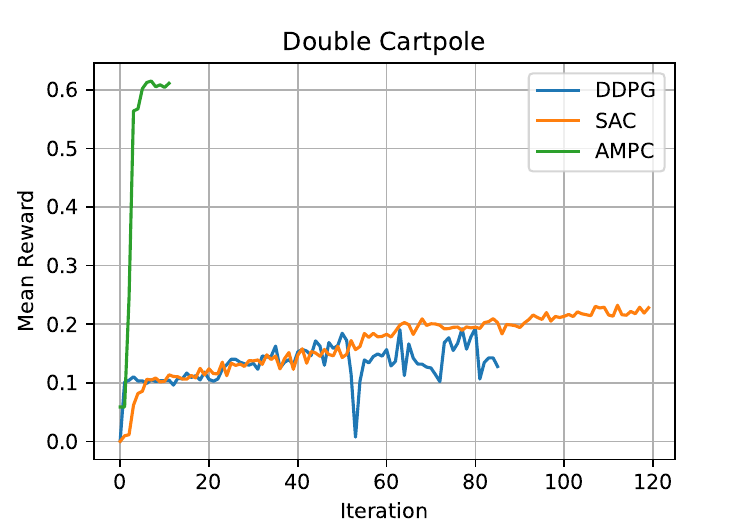}
    \caption{Evolution of the mean reward per training episode in the single (left) and double (right) cartpole environments of the model-free reinforcement learning algorithms SAC and DDPG, and our method, adaptive model-predictive control (AMPC).}
    \label{fig:ampc_rewards}
\end{figure}

Having a current model of the dynamics fitted to the true system dynamics, adaptive MPC significantly outperforms two model-free reinforcement learning baselines, Deep Deterministic Policy Gradient \cite{lillicrap2015continuous} and Soft Actor-Critic \cite{haarnoja2018soft}, in sample efficiency. Both baseline algorithms operate on the same state space as Adaptive MPC, while receiving a dense reward that matches the negative of our cost function. Although DDPG and SAC are able to eventually attain higher average rewards than adaptive MPC on the single cartpole swing-up task (Fig.~\ref{fig:ampc_rewards}), we note that the iLQR trajectory optimization constraints the force applied to the cartpole within a $[\SI{-200}{\newton}, \SI{200}{\newton}]$ interval, which caps the overall achievable reward as it takes more time to achieve the swing-up with less force-full movements.

\section{Related Work}
\label{sec:related}

Degrave et al.~\cite{degrave2019physics} implemented a differentiable physics engine in the automatic differentiation framework Theano. IDS is implemented in C++ using Stan Math~\cite{carpenter2015stan} which enables reverse-mode automatic differentiation to efficiently compute gradients, even in cases where the code branches significantly. Analytical gradients of rigid-body dynamics algorithms have been implemented efficiently in the Pinnocchio library~\cite{carpentier2018analytical} to facilitate optimal control and inverse kinematics. These are less general than our approach since they can only be used to optimize for a number of hand-engineered quantities. Simulating non-penetrative multi-point contacts between rigid bodies requires solving a linear complementarity problem (LCP), through which~\cite{peres2018lcp} differentiate using the differentiable quadratic program solver OptNet~\cite{amos2017optnet}. While our proposed model does not yet incorporate contact dynamics, we are able to demonstrate the scalability of our approach on versatile applications of differentiable physics to common 3D control domains.

Learning dynamics models has a tradition in the field of robotics and control theory. Early works on forward models~\cite{moore1992forward} and locally weighted regression~\cite{atkeson1997lwl} yielded control algorithms that learn from previous experiences. More recently, a variety of novel deep learning architectures have been proposed to learn \emph{intuitive physics} models. Inductive bias has been introduced through graph neural networks~\cite{hadsell2018graphnet, li2018learning, liu2019psd}, particularly interaction networks~\cite{battaglia2016interaction, schenck2018spnet, mrowca2018flexible} that are able to learn rigid and soft body dynamics. By incorporating more structure into the learning problem, Deep Lagrangian Networks~\cite{lutter2018delan} represent functions in the Lagrangian mechanics framework using deep neural networks. Besides novel architectures, vision-based machine learning approaches to predict the future outcomes of the state of the world have been proposed~\cite{wu2015galileo, wu2017deanimation, finn2016unsupervised}.

The approach of adapting the simulator to real world dynamics, which we demonstrate through our adaptive MPC algorithm in Sec.~\ref{sec:ampc}, has been less explored. While many previous works have shown to adapt simulators to the real world using system identification and state estimation~\cite{kolev2015sysid, zhu2018fastmi}, few have shown adaptive model-based control schemes that actively close the feedback loop between the real and the simulated system~\cite{reichenbach2009dynsim, farchy2013simback, chebotar2018sim2real}. Instead of using a simulator, model-based reinforcement learning is a broader field~\cite{polydoros2017mbrl}, where the system dynamics, and state-action transitions in particular, are learned to achieve higher sample efficiency compared to model-free methods. Within this framework, predominantly Gaussian Processes~\cite{ko2007gp, deisenroth2011pilco, boedecker2014sgp} and neural networks~\cite{williams2016mppi, yamaguchi2016neural} have been proposed to learn the dynamics and optimize policies.

\section{Future Work}

We plan to continue this contribution in several ways. IDS can only model limited dynamics due to its lack of a contact model. Modeling collision and contact in a plausibly differentiable way is an exciting topic that will greatly expand the number of environments that can be modeled. 

We are interested in exploring the loss surfaces of redundant physical parameters in IDS, where different models may have equivalent predictive power over the given task horizon. Resolving couplings between physical parameters can give rise to exploration strategies that expose properties of the physical system which allow our model to systematically calibrate itself. By examining the generalizability of models on these manifolds, we hope to establish guarantees of performance and prediction for specific tasks.

\section{Conclusion}
\label{sec:conclusion}

We introduced interactive differentiable simulation (IDS), a novel differentiable layer in the deep learning toolbox that allows for inference of physical parameters, optimal control and system design. Being constrained to the laws of physics, such as conservation of energy and momentum, our proposed model is interpretable in that its parameters have physical meaning. Combined with established learning algorithms from computer vision and receding horizon planning, we have shown how such a physics model can lead to significant improvements in sample efficiency and generalizability. Within a handful of trials in the test environment, our gradient-based representation of rigid-body dynamics allows an adaptive MPC scheme to infer the model parameters of the system thereby allowing it to make predictions and plan for actions many time steps ahead.

\medskip

\small

\bibliographystyle{plainnat}
\bibliography{literature}

\end{document}